%% file: arxiv.tex
\documentclass[10pt,twocolumn,letterpaper]{article}

\usepackage{cvpr}              % To produce the CAMERA-READY version
\input{preamble}

\definecolor{cvprblue}{rgb}{0.21,0.49,0.74}
\usepackage[pagebackref,breaklinks,colorlinks,citecolor=cvprblue]{hyperref}
\usepackage{epsfig}
\usepackage{graphicx}
\usepackage{amsmath}
\usepackage{amssymb}
\usepackage{multirow}
\usepackage{makecell} 
\usepackage{pifont}
\usepackage[ruled,linesnumbered]{algorithm2e}
\usepackage{algpseudocode}
\usepackage{verbatim}
\usepackage{tabularx}
\usepackage{array}
\usepackage{xcolor,colortbl}
\definecolor{Gray}{gray}{0.9}
\definecolor{LightCyan}{rgb}{0.88,1,1}
\newcolumntype{C}[1]{>{\centering\arraybackslash}p{#1}}

\def\paperID{1988} % *** Enter the Paper ID here
\def\confName{CVPR}
\def\confYear{2025}

\title{Self-Ensembling Gaussian Splatting for Few-Shot Novel View Synthesis}

\author{
Chen Zhao$^{1}$ \quad Xuan Wang$^{2}$ \quad Tong Zhang$^{1}$ \quad Saqib Javed$^{1}$ \quad Mathieu Salzmann$^{13}$ \\
$^1$EPFL \quad $^2$Ant Group  \quad $^3$Swiss Data Science Center \\
{\tt\small chen.zhao@epfl.ch} 
}

\begin{document}

\twocolumn[{
\renewcommand\twocolumn[1][]{#1}
\maketitle
\begin{center}
\vspace{-6mm}
\setlength{\abovecaptionskip}{4pt}
\centering
\includegraphics[width=1.0\linewidth]{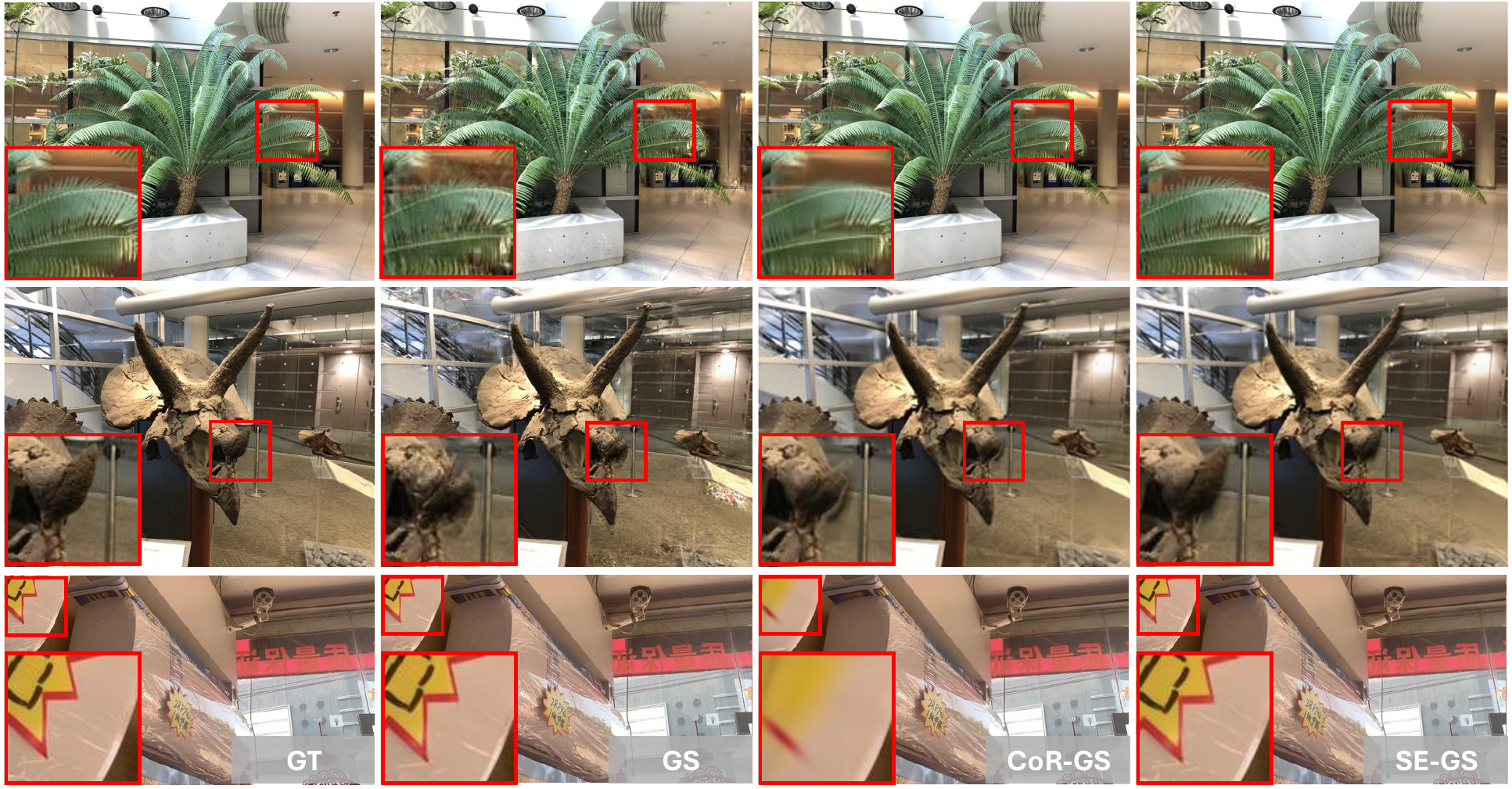}
\captionof{figure}{\textbf{Qualitative results of our SE-GS and state-of-the-art approaches.} The models are trained on sparse views and the images rendered from novel views are shown. As highlighted in the zoomed-in patches, our SE-GS captures finer details and produces fewer artifacts for novel views when trained on few-shot images.} 
\label{fig:demo1} 
\end{center}
}]

\input{sec/0_abstract_arxiv}    
\input{sec/1_intro}
\input{sec/2_related_work}
\input{sec/3_method}
\input{sec/4_experiments}
\input{sec/5_conclusion_arxiv}
\input{sec/X_suppl}

{
    \bibliographystyle{ieeenat_fullname}
    \bibliography{main}
}


\end{document}

%% file: preamble.tex
\usepackage[dvipsnames]{xcolor}

%% file: sec/0_abstract_arxiv.tex
\begin{abstract}
3D Gaussian Splatting (3DGS) has demonstrated remarkable effectiveness in novel view synthesis (NVS). However, 3DGS tends to overfit when trained with sparse views, limiting its generalization to novel viewpoints. In this paper, we address this overfitting issue by introducing Self-Ensembling Gaussian Splatting (SE-GS). We achieve self-ensembling by incorporating an uncertainty-aware perturbation strategy during training. A $\mathbf{\Delta}$-model and a $\mathbf{\Sigma}$-model are jointly trained on the available images. The $\mathbf{\Delta}$-model is dynamically perturbed based on rendering uncertainty across training steps, generating diverse perturbed models with negligible computational overhead. Discrepancies between the $\mathbf{\Sigma}$-model and these perturbed models are minimized throughout training, forming a robust ensemble of 3DGS models. This ensemble, represented by the $\mathbf{\Sigma}$-model, is then used to generate novel-view images during inference. Experimental results on the LLFF, Mip-NeRF360, DTU, and MVImgNet datasets demonstrate that our approach enhances NVS quality under few-shot training conditions, outperforming existing state-of-the-art methods. The code is released at: \href{https://sailor-z.github.io/projects/SEGS.html}{project page}.
\end{abstract}

%% file: sec/1_intro.tex
\section{Introduction}
\label{sec:intro}
\begin{figure}[!t]
    \centering	
    \begin{subfigure}{0.49\linewidth}
        \includegraphics[width=1.0\linewidth]{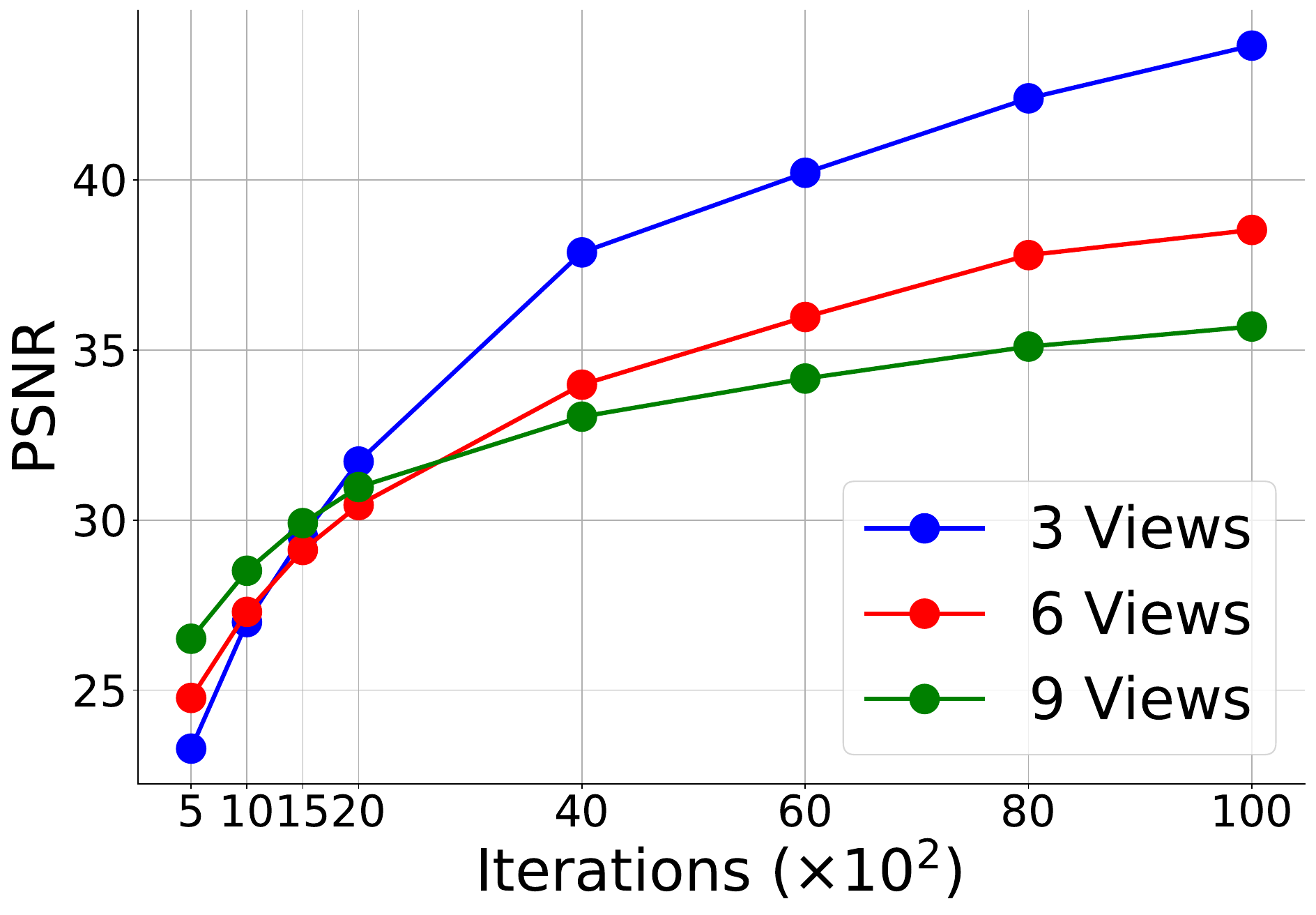}
        \caption{Training}
        \label{fig:intro_a}
    \end{subfigure}
    \hfill
    \begin{subfigure}{0.49\linewidth}
        \includegraphics[width=1.0\linewidth]{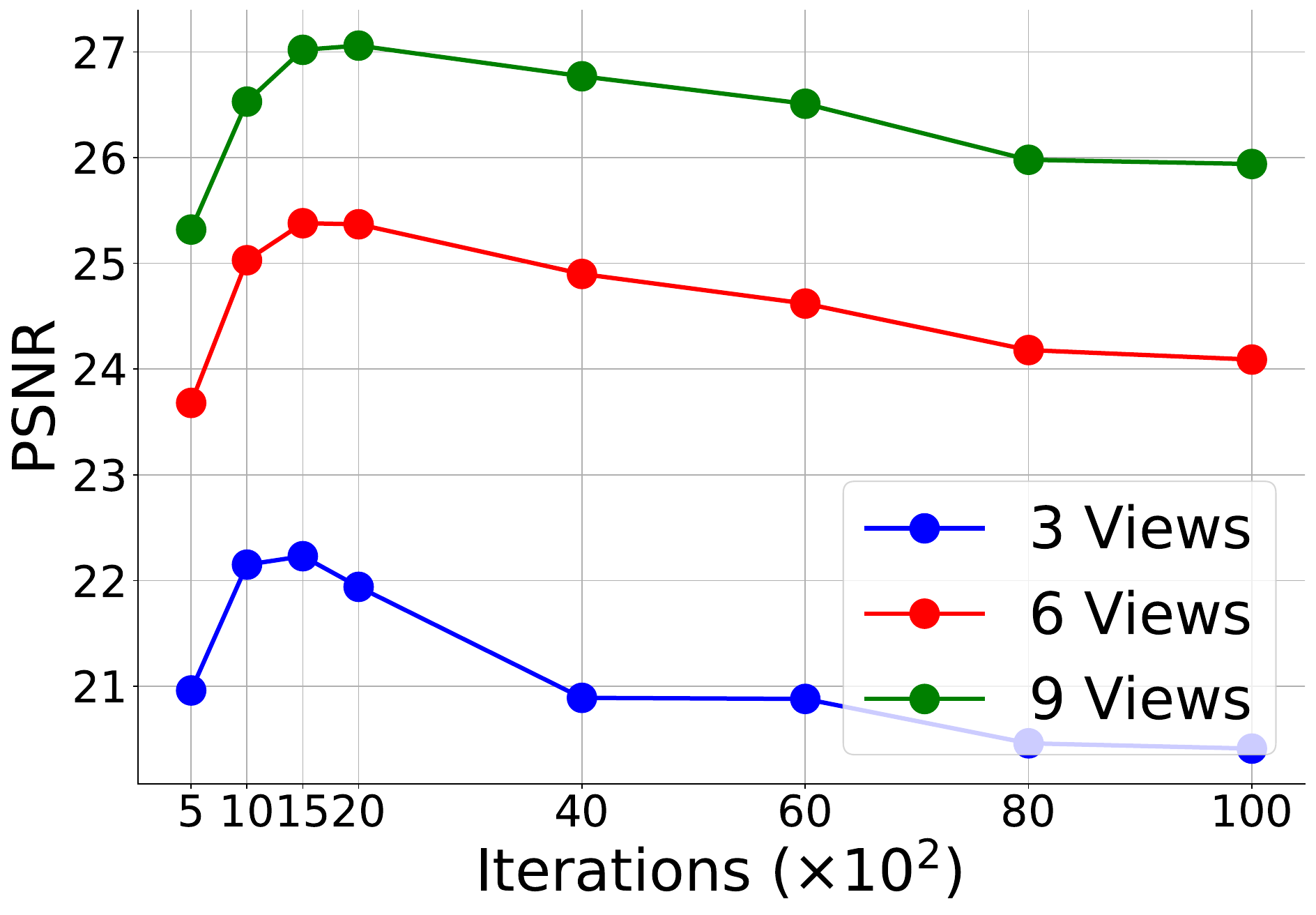}
        \caption{Testing}
    \label{fig:intro_b}
\end{subfigure}
    \caption{\textbf{Overfitting in 3D Gaussian Splatting with sparse training views.} (a) and (b) illustrate the performance of 3DGS on training and testing views, respectively. Each curve represents the PSNR values across training iterations.} 
    \label{fig:intro}
    \vspace{-10pt}
\end{figure}

Novel view synthesis (NVS) is a critical task~\cite{zhou2016view} in computer vision and graphics, playing a pivotal role in applications such as virtual reality~\cite{cho2019novel}, augmented reality~\cite{zhou2018stereo}, and 3D content generation~\cite{guedon2024sugar,wolf2024surface}. The objective of NVS is to generate photo-realistic images from previously unseen viewpoints. Typically, NVS starts by constructing a 3D representation~\cite{mildenhall2021nerf,wang2021neus} from a set of existing 2D observations. In recent years, 3D Gaussian Splatting (3DGS)~\cite{kerbl20233d,chen2024survey,yu2024mip} has emerged as a powerful representation, integrating the advantages of both explicit~\cite{schoenberger2016sfm} and implicit~\cite{mildenhall2021nerf} representations. This approach enables efficient novel view generation and yields promising synthesized results with densely sampled observations that cover a wide range of viewpoints. However, 3DGS tends to overfit the available views when only a limited number of images are provided. To illustrate this issue, in Fig.~\ref{fig:intro}, we evaluate 3DGS trained on sparse images with different numbers of iterations. The performance on the training data consistently improves as the number of iterations increases, while the testing results deteriorate after 2000 iterations. Moreover, the overfitting problem becomes more noticeable with fewer training views, such as when using only 3 views. 

To mitigate overfitting, ensembling~\cite{lakshminarayanan2017simple,laine2016temporal,fort2019deep} has been highlighted as an effective strategy in detection~\cite{zheng2021se} and segmentation~\cite{li2020transformation}. Nevertheless, its use for NVS remains unstudied, and how to exploit it in a 3DGS formalism is an open question. Therefore, in this paper, we bridge this gap, introducing a new 3DGS method that enhances the quality of novel view synthesis with sparse training views via \emph{self-ensembling}. A straightforward way to achieve ensembling would be to train multiple 3DGS models and combine the corresponding predictions as the final result during testing. However, as we will demonstrate in Sec.~\ref{sec:exp}, this method is computationally expensive, and the resulting models lack sufficient diversity to ensure effective ensembling. 

In contrast, we present an effective and efficient self-ensembling mechanism. In ensemble learning~\cite{french2017self,nguyen2019self}, self-ensembling is typically achieved by introducing perturbations during training through augmentation or dropout. Inspired by its success, we introduce an uncertainty-aware perturbation strategy for 3DGS. Specifically, we train a $\mathbf{\Delta}$-model on available RGB images. During training, we perturb the $\mathbf{\Delta}$-model based on uncertainties derived from the renderings. To compute these uncertainties, we store images rendered from the $\mathbf{\Delta}$-model at pseudo views across different training iterations in buffers and calculate pixel-level uncertainties within each buffer. A perturbed model is then obtained by adding random noise to the Gaussian parameters of the $\mathbf{\Delta}$-model associated with pixels that have high uncertainty scores. Since all perturbed models are derived from the $\mathbf{\Delta}$-model rather than being trained from scratch, our perturbation strategy generates diverse models without incurring significant computational overhead. In addition to the $\mathbf{\Delta}$-model, we train a $\mathbf{\Sigma}$-model on the training views without the perturbation. The self-ensembling is achieved by minimizing the discrepancies between the $\mathbf{\Sigma}$-model and perturbed models. These discrepancies are measured via a photometric loss between images synthesized at the pseudo views. This self-ensembling is performed based on diverse models, thereby improving the robustness and generalization of the resulting ensemble, the $\mathbf{\Sigma}$-model. Moreover, our self-ensembling is independent of any additional ground-truth signals as it is carried out in a self-supervised manner. During testing, the $\mathbf{\Sigma}$-model is employed for novel view synthesis.

We conduct experiments on multiple datasets, including LLFF~\cite{mildenhall2019local}, DTU~\cite{jensen2014large}, Mip-NeRF360~\cite{barron2022mip}, and MVImgNet~\cite{yu2023mvimgnet}, with sparse training views. Our SE-GS achieves the best performance across all datasets in the sparse-view setting, surpassing the state-of-the-art approaches. Furthermore, we perform a comprehensive analysis of our method to demonstrate its effectiveness and efficiency. To the best of our knowledge, we are the first to explore the potential of the self-ensembling mechanism in 3DGS for few-shot novel view synthesis. The code will be publicly available upon acceptance.

%% file: sec/2_related_work.tex
\section{Related Work}
\label{sec:related_work}
\noindent{\textbf{Radiance field modeling.}} Recently, Neural Radiance Field (NeRF)~\cite{mildenhall2021nerf,wang2021nerf,pumarola2021d,barron2022mip} has brought groundbreaking advancements to novel view synthesis. However, most NeRF-based methods are inefficient~\cite{hu2022efficientnerf} because they require numerous MLP queries in the rendering process. This limitation makes it challenging to support tasks with real-time requirements. 
In this context, 3D Gaussian Splatting~\citep{kerbl20233d} is developed as an efficient alternative for radiance field modeling. 3DGS relies on explicit representations~\cite{schoenberger2016sfm} that allow for faster rendering and training~\cite{fan2024instantsplat}. While initially introduced as a scene-specific model, some 3DGS variants~\cite{wang2024freesplat,charatan2024pixelsplat,liu2025mvsgaussian,chen2024mvsplat} explore the generalization of 3DGS towards novel scenes, optimizing the Gaussian parameters based on learnable cost volumes~\cite{gu2020cascade}. In this paper, we stick to the vanilla 3DGS rendering pipeline, focusing on the scene-level radiance field modeling.

\noindent{\textbf{3DGS with few-shot images.}} The insufficiency of training views is a primary factor leading to artifacts in novel view synthesis. Even when training views are relatively dense, some regions may still be observed from few-shot images, resulting in quality degradation in the synthesized views. To handle such an under-constrained problem, some approaches incorporate additional ground-truth signals into the 3DGS pipeline. For instance, DNGaussian~\cite{li2024dngaussian}, CoherentGS~\cite{paliwal2024coherentgs}, and FSGS~\cite{zhu2024fsgs} utilize a monocular depth estimator~\cite{ranftl2021vision} to predict depth maps, enabling the 3DGS model to be trained with both a photometric and a depth loss. Other methods, such as~\cite{xu2024mvpgs} and~\cite{han2024binocular}, leverage multi-view stereo techniques~\cite{seitz2006comparison} to generate novel-view images as ground truth. However, ground-truth data obtained from off-the-shelf methods is inevitably noisy, which potentially impacts the training of 3DGS. As noted in~\cite{zhang2024cor} and observed in our experiments, when the number of training views increases, the performance of these methods is sometimes worse than the vanilla 3DGS. To overcome this problem, Zhang~\emph{et al.}~\cite{zhang2024cor} propose training multiple 3DGS models with a cross-model regularization term. However, training additional 3DGS models incurs a significant computational cost, making it impractical to scale up to a large number of 3DGS models for stronger regularization.

\noindent{\textbf{Ensemble learning.}} Ensembling has been evidenced as a powerful technique in machine learning~\cite{zhou2002ensembling,ashukha2020pitfalls,garipov2018loss} to improve model robustness and generalization by aggregating predictions from multiple models. Traditionally, ensembles~\cite{ganaie2022ensemble,rahaman2021uncertainty,lee2015m} are created by training independent models and averaging their outputs or applying a voting mechanism. To improve the efficiency, some self-ensembling methods, such as temporal ensembling~\cite{laine2016temporal} and consistency regularization~\cite{french2017self}, leverage variations of a single model across training iterations to build an ensemble. Motivated by the success of self-ensembling, we present the first Self-Ensembling Gaussian Splatting approach, in which we generate diverse samples in the Gaussian parameter space through an uncertainty-aware perturbation mechanism. 

%% file: sec/3_method.tex
\section{Method}
\label{sec:method}

\begin{figure*}[!t]
    \includegraphics[width=1.0\linewidth]{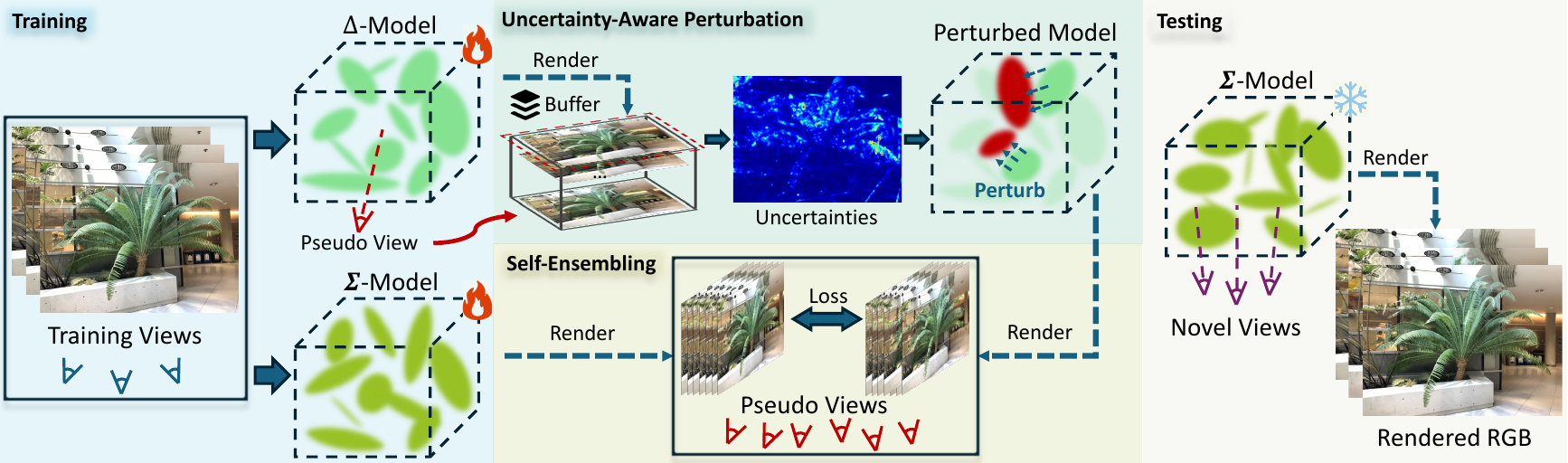}
    \caption{\textbf{Pipeline of the presented SE-GS.} We tackle the overfitting problem in sparse-view scenarios by incorporating a self-ensembling mechanism into 3DGS. We jointly train a $\mathbf{\Delta}$-model and a $\mathbf{\Sigma}$-model. During training, we store pseudo-view renderings of the $\mathbf{\Delta}$-model in buffers, from which we compute pixel-level uncertainties. The Gaussians of the $\mathbf{\Delta}$-model overlapping the pixels with high uncertainties are perturbed, as highlighted as red ellipses, which leads to a perturbed model. We then achieve self-ensembling by penalizing the discrepancies between the $\mathbf{\Sigma}$-model and the perturbed models. During inference, the resulting ensemble, the $\mathbf{\Sigma}$-model, is used for novel view synthesis.}
    \label{fig:pipeline}
\end{figure*}

\subsection{Preliminaries}
3D Gaussian Splatting~\cite{kerbl20233d} represents a scene as a collection of Gaussians. These Gaussians are splatted onto the 2D image plane during rendering. Formally, we denote the set of $N$ Gaussians as $\{{G_i}, i=1,2,...,N\}$. Each Gaussian $G_i$ is defined as $G_i =(\mathbf{\mu}_i, \Sigma_i, \mathbf{h}_i, o_i)$, where $\mathbf{\mu}_i$ is the 3D position, $\Sigma_i$ is the covariance matrix, $\mathbf{h}_i$ represents spherical harmonics (SH) coefficients associated with the Gaussian, and $o_i$ indicates opacity. Each Gaussian contributes to a 3D point $\mathbf{x}$ according to the 3D Gaussian distribution
\begin{align}
\label{eq:3dgs}
\mathcal{N}_{3d}^{i}(\mathbf{x}) = e^{-\frac{1}{2}(\mathbf{x} - \mathbf{\mu}_{i})^T \Sigma_{i}^{-1} (\mathbf{x} - \mathbf{\mu}_{i})}.
\end{align}
To ensure that $\Sigma_{i}$ remains positive semi-definite throughout optimization, it is decomposed into two learnable components as $\Sigma_{i}=\mathbf{R}\mathbf{S}\mathbf{S}^{T}\mathbf{R}^{T}$, where $\mathbf{R}$ is a rotation matrix and $\mathbf{S}$ stands for a scaling matrix. For RGB rendering, the projection from 3D space to a 2D image plane relies on the projection matrix $\mathbf{W}$ to compute the projected 2D covariance matrix
\begin{align}
\label{eq:proj}
\Sigma_{i}^{'}=\mathbf{J}\mathbf{W}\Sigma_{i}\mathbf{W}^{T}\mathbf{J}^{T},
\end{align}
where $\mathbf{J}$ denotes the Jacobian of the affine approximation of the projection. The color of a pixel is obtained by performing alpha-blending as
\begin{align}
\label{eq:render}
\mathbf{c} = \sum_{i=1}^{M} \mathbf{c}_i \alpha_i \prod_{j=1}^{i-1} (1 - \alpha_j),
\end{align}
where $M$ is the number of Gaussians covering the pixel, $\mathbf{c}_i$ denotes the color of the Gaussian derived from the SH coefficients, and $\alpha_{i}$ is computed from the 2D covariance matrices and opacity scores. For better convergence, the Gaussian parameters are initialized based on 3D points obtained using structure-from-motion techniques~\cite{schoenberger2016sfm,schoenberger2016mvs}. These parameters are optimized with a photometric loss~\cite{kerbl20233d} where the posed training images serve as ground truth.

\subsection{Motivations}
In this paper, we focus on sparse-view scenarios, where few-shot images are provided. We denote the 3DGS model trained on these images as $\mathcal{G}$. In this setting, $\mathcal{G}$ is prone to overfitting the training data and thereby getting trapped in a suboptimal solution, which ultimately degrades the quality of the synthesized novel views. A promising approach to mitigate overfitting is ensemble learning~\cite{ganaie2022ensemble}, which has been utilized to enhance robustness and generalization in the literature. Rather than depending on a single model, ensemble learning aggregates predictions from multiple models, thereby stabilizing the final output. To achieve this, a straightforward method is to jointly train a set of 3DGS models $\{\mathcal{G}_1, \mathcal{G}_2,..., \mathcal{G}_k\}$. As we will report in Sec.~\ref{sec:exp}, by aggregating their predictions, the final results are more robust, and the NVS performance improves as $k$ grows. However, training multiple 3DGS models incurs significant computational costs, making it impractical for large $k$. Moreover, our experimental findings indicate that the trained 3DGS models are not diverse enough to support effective ensembling. Consequently, we propose a novel self-ensembling paradigm based on an uncertainty-aware perturbation strategy, enhancing 3DGS for few-shot NVS with negligible additional training overhead. The pipeline is shown in Fig.~\ref{fig:pipeline} and the details will be elaborated in this section.
 
\subsection{Uncertainty-Aware Perturbation} 
\begin{figure}[!t]
    \includegraphics[width=1.0\linewidth]{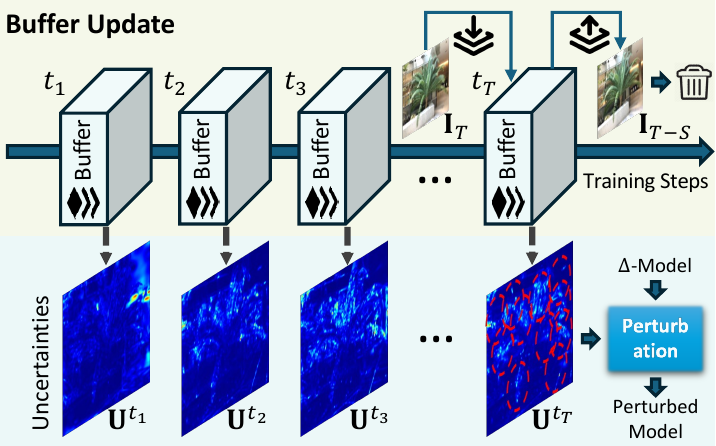}
    \caption{\textbf{Buffer update during training.} For each sampled pseudo view, we dynamically update the buffer storing the images rendered at different training steps. For instance, at training step $t_{T}$, the oldest image $\mathbf{I}_{T-S}$ in the buffer is popped, and the new image $\mathbf{I}_{T}$ is pushed into the buffer. An uncertainty map $\mathbf{U}^{t_{T}}$ is computed based on the current buffer, which is then employed to determine perturbation that results in a new 3DGS model.}
    \label{fig:temporal}
\end{figure}

In contrast to training separate 3DGS models from scratch, we dynamically generate diverse models from a single $\mathbf{\Delta}$-model during training. Specifically, we train the $\mathbf{\Delta}$-model on the available posed images, following the same optimization and density control strategies introduced in 3DGS~\cite{kerbl20233d}. At each training iteration, the current $\mathbf{\Delta}$-model represents a specific sample in the Gaussian parameter space based on the information contained in the training data. A new 3DGS model is then created by perturbing the $\mathbf{\Delta}$-model as
\begin{align}
\label{eq:perturb}
\hat{G}_{\Delta}^{t}=G_{\Delta}^{t}+\delta_{t},
\end{align}
where $G_{\Delta}^{t}$ stands for a Gaussian in the $\mathbf{\Delta}$-model at training step $t$, $\delta_{t}\in \mathcal{N} (\mathbf{\mu_{t}}, \mathbf{\sigma}^{2}_t)$ indicates the noise, and $\hat{G}_{\Delta}^{t}$ denotes a perturbed Gaussian in the perturbed model. This naive approach adds random noise to all Gaussians in the $\mathbf{\Delta}$-model. However, as we will demonstrate in Sec.~\ref{sec:exp}, this method shifts perturbed models too far from the $\mathbf{\Delta}$-model, leading to instability and unreliable supervision.

Therefore, we present an uncertainty-aware perturbation strategy, leveraging the statistics of renderings. Our approach starts by creating $M$ pseudo views through interpolation between the training views. Specifically, given two cameras sampled from the training views, the camera extrinsics of a pseudo view are computed as
\begin{align}
\label{eq:pseudo}
& \hat{\mathbf{R}}=\text{SLERP}(\mathbf{R}_1, \mathbf{R}_2, \beta), \\
& \hat{\mathbf{c}}= \beta\mathbf{c}_1 + (1-\beta)\mathbf{c}_2, \\
& \hat{\mathbf{T}} = -\hat{\mathbf{R}}\hat{\mathbf{c}},
\end{align}
where $(\mathbf{R}_1, \mathbf{R}_2)$ represent the rotations of the sampled cameras, $(\mathbf{c}_1, \mathbf{c}_2)$ indicate the sampled camera centers, SLERP denotes the spherical linear interpolation~\cite{buss2001spherical}, $\beta$ is a randomly sampled scalar that controls the interpolation, and $(\hat{\mathbf{R}}, \hat{\mathbf{T}})$ stand for the camera parameters of the pseudo view. As illustrated in Fig.~\ref{fig:pipeline}, for each pseudo view, we render an RGB image using the current $\mathbf{\Delta}$-model and store the renderings at different training steps in a buffer. We then compute a pixel-wise uncertainty map
\begin{align}
\label{eq:uncerntainty}
\mathbf{U}=\sqrt{\frac{1}{S}\sum_{i=1}^{S}(\mathbf{I}_i-\bar{\mathbf{I}})^{2}},
\end{align}
where $\mathbf{I}_i$ represents an image in the buffer, $\bar{\mathbf{I}}$ indicates the mean of these images, and $S$ is the buffer size. This uncertainty estimation is carried out over all $M$ sampled pseudo views in parallel. To enhance robustness, we perform local smoothing over each uncertainty map, yielding
\begin{align}
\label{eq:smooth}
\hat{\mathbf{U}}(i, j) = \frac{1}{k^{2}}\sum_{m,n} \mathbf{U}(m, n),
\end{align}
where $k=5$ is the size of a kernel centered at $(i, j)$ and $(m, n)$ indicate the coordinates of a pixel within the kernel. Subsequently, we apply uncertainty-aware perturbation to each Gaussian in the $\mathbf{\Delta}$-model as
\begin{align}
\label{eq:perturb_ours}
\hat{G}_{\Delta}^{t}=G_{\Delta}^{t}+\delta_{t}h(G_{\Delta}^{t}, \hat{\mathcal{U}}^{t}),
\end{align}
where $\hat{\mathcal{U}}^{t}=\{\hat{\mathbf{U}}^t_1, \hat{\mathbf{U}}^t_2, \cdots, \hat{\mathbf{U}}^t_M\}$ is a set of uncertainty maps computed from the buffers at the current training step, and $h(\cdot, \cdot)$ is an indicator function defined as
\begin{align} 
\label{eq:indicator} 
h(G_{\Delta}^{t}, \hat{\mathcal{U}}^{t}) = 
\begin{cases} 
1 & \text{if } \max\limits_{(i,j)\in\mathcal{P}(G_{\Delta}^{t})}u_{ij} \geq \tau \, \\
0 & \text{if } \max\limits_{(i,j)\in\mathcal{P}(G_{\Delta}^{t})}u_{ij} < \tau,
\end{cases} 
\end{align}
with $\tau$ a predefined threshold, and $u_{ij}$ the uncertainty score of a pixel in a set of pixels $\mathcal{P}$ overlapping with the 2D splats of $G_{\Delta}^{t}$. Please refer to the supplementary material for details on defining $\tau$ and identifying $\mathcal{P}$. In short, the reliability of each Gaussian is connected with the rendering uncertainty, and only the unreliable Gaussians, characterized by high uncertainty scores, are perturbed. In practice, we add random noise to 3D positions, 3D rotations, scales, and opacities of the $\mathbf{\Delta}$-model. Particularly, since rotation is not continuous when expressed as a 3D matrix, we perturb the 6D continuous representation~\cite{zhou2019continuity}, which is denoted as 
\begin{align} 
\label{eq:rota}
\hat{\mathbf{R}}^{t}_{\Delta}=f^{-1}(f(\mathbf{R}_{\Delta}^{t}) + \delta^{R}_{t}h(G_{\Delta}^{t}, \hat{\mathcal{U}}^{t})), \ \ \delta^{R}_{t}\in\mathbb{R}^{6},
\end{align}
where $f(\cdot)$ indicates the mapping from rotation matrix to 6D representation.
Fig.~\ref{fig:temporal} illustrates the dynamic updates of the buffer throughout the training process. As the buffer is updated, the uncertainty map varies accordingly, resulting in diverse perturbed models from the $\mathbf{\Delta}$-model.

It is worth noting that the number of Gaussians in the $\mathbf{\Delta}$-model varies during training due to density control, making it challenging to assess uncertainties directly in the Gaussian parameter space for newly generated Gaussians. Our approach naturally handles this challenge as the statistics on 2D renderings provide a consistent ground to measure uncertainties, regardless of the varying number of Gaussians.

\subsection{Self-Ensembling in 3DGS} 
Given the perturbed models derived from the $\mathbf{\Delta}$-model, the next step in our pipeline is to construct an ensemble from these models. To ensure efficiency, we perform self-ensembling during training. Concretely, we train a separate 3DGS model, named the $\mathbf{\Sigma}$-model, on the training views without perturbation. Its training is guided with an additional regularization formulated as
\begin{align}
\label{eq:reg}
\mathcal{L}_{r}=(1-\lambda){||\mathbf{I}^{t}_{\Sigma} - \mathbf{I}^{t}_{\Delta}||}_1+\lambda\mathcal{L}_{\text{D-SSIM}}(\mathbf{I}^{t}_{\Sigma},\mathbf{I}^{t}_{\Delta}),
\end{align}
where $\lambda=0.2$ is a predefined weight, $\mathcal{L}_{\text{D-SSIM}}$ denotes a D-SSIM term~\cite{kerbl20233d}, and $(\mathbf{I}^{t}_{\Sigma},\mathbf{I}^{t}_{\Delta})$ represent the images rendered from a pseudo view using the current $\mathbf{\Sigma}$-model and the perturbed model, respectively. We also utilize a co-pruning strategy~\cite{zhang2024cor} to enhance the regularization. Therefore, the $\mathbf{\Sigma}$-model aggregates information from diverse perturbed models, functioning as an ensemble of 3DGS models. The final loss function during training is defined as
\begin{align}
\label{eq:loss}
\mathcal{L}=\mathcal{L}_{\text{RGB}}+\gamma\mathcal{L}_{r},
\end{align}
where $\gamma=1$ by default and $\mathcal{L}_{\text{RGB}}$ represents a photometric loss over the training views. During testing, we keep the $\mathbf{\Sigma}$-model for novel view synthesis. Notably, compared with the previous approach~\cite{zhang2024cor}, our SE-GS inherently encodes diverse 3DGS models without significant computational overhead, thereby enabling efficient and effective regularization. Moreover, the regularization is applied on pseudo views in a self-supervised manner, making it independent of additional information such as depth~\cite{li2024dngaussian,zhu2024fsgs}. 

\begin{table*}[!t]   
  \centering
  \begin{tabular}{@{}l|ccc|ccc|ccc}
    \toprule
     \multirow{2}{*}{Method}   & \multicolumn{3}{c|}{PSNR$\uparrow$}  & \multicolumn{3}{c|}{SSIM$\uparrow$}   & \multicolumn{3}{c}{LPIPS$\downarrow$} \\
      &  3-view & 6-view & 9-view & 3-view & 6-view & 9-view & 3-view & 6-view & 9-view \\
    \midrule
    Mip-NeRF~\cite{barron2021mip} & 16.11 & 22.91 & 24.88 & 0.401 & 0.756 & 0.826 & 0.460 & 0.213 & 0.160 \\
    DietNeRF~\cite{jain2021putting} & 14.94 & 21.75 & 24.28 & 0.370 & 0.717 & 0.801 & 0.496 & 0.248 & 0.183 \\
    RegNeRF~\cite{niemeyer2022regnerf} & 19.08 & 23.10 & 24.86 & 0.587 & 0.760 & 0.820 & 0.336 & 0.206 & 0.161 \\
    FreeNeRF~\cite{yang2023freenerf} & 19.63 & 23.73 & 25.13 & 0.612 & 0.779 & 0.827 & 0.308 & 0.195 & 0.160 \\
    SparseNeRF~\cite{wang2023sparsenerf} & 19.86 & - & - & 0.624 & - & - & 0.328 & - & - \\
    \hline
    3DGS~\cite{kerbl20233d} & 19.22 & 23.80 & \cellcolor[HTML]{FFFFD4}25.44 & 0.649 & 0.814 & \cellcolor[HTML]{FFFFD4}0.860 & \cellcolor[HTML]{FFFFD4}0.229 & \cellcolor[HTML]{FFFFD4}0.125 & \cellcolor[HTML]{FFFFD4}0.096 \\
    DNGaussian~\cite{li2024dngaussian} & 19.12 & 22.01 & 22.62 & 0.591 & 0.717 & 0.741 & 0.294 & 0.246 & 0.244 \\
    % MVPGS~\cite{xu2024mvpgs} & \cellcolor[HTML]{FFFFD4}20.54 & \cellcolor[HTML]{FFFFD4}23.68 & 24.14 & \cellcolor[HTML]{FFFFD4}0.727 & \cellcolor[HTML]{FFFFD4}0.819 & \cellcolor[HTML]{FFFFD4}0.843 & 0.194 & 0.136 & 0.120 & - & - & - \\
    FSGS~\cite{zhu2024fsgs} & \cellcolor[HTML]{FFFFD4}20.43 & \cellcolor[HTML]{FFFFD4}24.09 & 25.31 & \cellcolor[HTML]{FFFFD4}0.682 & \cellcolor[HTML]{FFFFD4}0.823 & \cellcolor[HTML]{FFFFD4}0.860 & 0.248 & 0.145 & 0.122  \\
    CoR-GS~\cite{zhang2024cor} & \cellcolor[HTML]{FFE4CF}20.45 & \cellcolor[HTML]{FFE4CF}24.49 & \cellcolor[HTML]{FFE4CF}26.06 & \cellcolor[HTML]{FFE4CF}0.712 & \cellcolor[HTML]{FFE4CF}0.837 & \cellcolor[HTML]{FFE4CF}0.874 & \cellcolor[HTML]{FFE4CF}0.196 & \cellcolor[HTML]{FFE4CF}0.115 & \cellcolor[HTML]{FFE4CF}0.089 \\
    \textbf{SE-GS} & \cellcolor[HTML]{FFCCC9}20.79 & \cellcolor[HTML]{FFCCC9}24.78 & \cellcolor[HTML]{FFCCC9}26.36 & \cellcolor[HTML]{FFCCC9}0.724 & \cellcolor[HTML]{FFCCC9}0.839 & \cellcolor[HTML]{FFCCC9}0.878 & \cellcolor[HTML]{FFCCC9}0.183 & \cellcolor[HTML]{FFCCC9}0.110 & \cellcolor[HTML]{FFCCC9}0.084 \\
  \bottomrule
  \end{tabular}

  \caption{\textbf{Results on LLFF with 3, 6, and 9 training views.} We highlight the best, second-best, and third-best results in red, orange, and yellow, respectively.}
  \label{tab:llff}
\end{table*}

\begin{table*}[!t]
    \centering
    \begin{tabular}{l|ccc|ccc|ccc}
        \toprule
        \multirow{2}{*}{Method}   & \multicolumn{3}{c|}{PSNR$\uparrow$}  & \multicolumn{3}{c|}{SSIM$\uparrow$}   & \multicolumn{3}{c}{LPIPS$\downarrow$} \\
        &  3-view & 6-view & 9-view & 3-view & 6-view & 9-view & 3-view & 6-view & 9-view \\
        \midrule
        3DGS~\cite{kerbl20233d} & 17.67 & \cellcolor[HTML]{FFFFD4}23.69 & \cellcolor[HTML]{FFFFD4}26.80 & 0.804 & 0.894 & \cellcolor[HTML]{FFFFD4}0.941 & \cellcolor[HTML]{FFFFD4}0.158 & \cellcolor[HTML]{FFFFD4}0.086 & \cellcolor[HTML]{FFFFD4}0.050 \\
        MVSplat~\cite{chen2024mvsplat} & 17.33 & 16.34 & 16.22 & 0.598 & 0.596 & 0.587 & 0.279 & 0.296 & 0.304 \\
        DNGaussian~\cite{li2024dngaussian} & \cellcolor[HTML]{FFFFD4}18.57 & 22.56 & 25.25 & 0.776 & 0.862 & 0.917 & 0.178 & 0.114 & 0.077 \\
        FSGS~\cite{zhu2024fsgs} & 17.84 & 23.68 & 26.17 & \cellcolor[HTML]{FFFFD4}0.822 & \cellcolor[HTML]{FFFFD4}0.905 & \cellcolor[HTML]{FFFFD4}0.941 & 0.161 & 0.096 & 0.064 \\
        CoR-GS~\cite{zhang2024cor} & \cellcolor[HTML]{FFE4CF}18.65 & \cellcolor[HTML]{FFE4CF}24.39 & \cellcolor[HTML]{FFE4CF}27.38 & \cellcolor[HTML]{FFE4CF}0.835 & \cellcolor[HTML]{FFE4CF}0.910 & \cellcolor[HTML]{FFE4CF}0.950 & \cellcolor[HTML]{FFE4CF}0.140 & \cellcolor[HTML]{FFE4CF}0.074 & \cellcolor[HTML]{FFE4CF}0.045 \\
        \textbf{SE-GS} & \cellcolor[HTML]{FFCCC9}19.24 & \cellcolor[HTML]{FFCCC9}25.28 & \cellcolor[HTML]{FFCCC9}28.08 & \cellcolor[HTML]{FFCCC9}0.857 & \cellcolor[HTML]{FFCCC9}0.924 & \cellcolor[HTML]{FFCCC9}0.958 & \cellcolor[HTML]{FFCCC9}0.132 & \cellcolor[HTML]{FFCCC9}0.073 & \cellcolor[HTML]{FFCCC9}0.043 \\
        \bottomrule
    \end{tabular}
    \caption{\textbf{Results on DTU with 3, 6, and 9 training views.} We use red, orange, and yellow to indicate the best, second-best, and third-best results, respectively. Object masks are used for all evaluated methods to remove background when conducting the evaluation.}
    \label{tab:dtu}
\end{table*}

%% file: sec/4_experiments.tex
\section{Experiments}
\label{sec:exp}

\subsection{Setup}
\noindent{\textbf{Datasets.}} We conduct experiments on four datasets, i.e., LLFF~\cite{mildenhall2019local}, DTU~\cite{jensen2014large}, Mip-NeRF360~\cite{barron2022mip}, and MVImgNet~\cite{yu2023mvimgnet}. On LLFF, DTU, and Mip-NeRF360, we follow the experimental setup introduced in~\cite{zhang2024cor}, using the same training/testing splits. As suggested in~\cite{zhang2024cor}, we mask the background when assessing the quality of novel view synthesis on the DTU dataset. Since LLFF, DTU, and Mip-NeRF360 offer a limited number of scenarios, we extend our experiments to a large-scale dataset, i.e., MVImgNet. We randomly sample 50 scenes from this dataset and resize the longest side of each image to 512. Notably, in our experiments, the Gaussian parameters for all evaluated methods are initialized based on the same point clouds obtained from COLMAP~\cite{schoenberger2016mvs}, ensuring a fair comparison. Moreover, COLMAP fails on certain DTU scenes in the sparse-view setting. In these cases, we instead randomly initialize the point cloud. Please refer to the supplementary material for more details on the setup.

\noindent{\textbf{Implementation details.}} We train our SE-GS for 10,000 iterations on the LLFF, DTU, and MVImgNet datasets, and for 30,000 iterations on Mip-NeRF360. In our uncertainty-aware perturbation mechanism, we employ $M=24$ image buffers with a buffer size of $S=3$ and perturb the $\mathbf{\Delta}$-model every 500 iterations. The noise $\delta_t$ in Eq.~\ref{eq:perturb_ours} is sampled from a normal distribution with a mean of 0, and the standard deviation is adaptively adjusted based on the magnitude of the Gaussian parameters. More details are provided in the supplementary material.

\begin{table}[!t]   
  \centering
  \resizebox{1.0\linewidth}{!}{
  \begin{tabular}{@{}l|ccc|ccc}
    \toprule
     \multirow{2}{*}{Method}   & \multicolumn{3}{c|}{12-view}  & \multicolumn{3}{c}{24-view} \\
      & PSNR$\uparrow$ & SSIM$\uparrow$ & LPIPS$\downarrow$ & PSNR$\uparrow$ & SSIM$\uparrow$ & LPIPS$\downarrow$ \\
    \midrule
    3DGS~\cite{kerbl20233d} & 18.52 & 0.523 & \cellcolor[HTML]{FFE4CF}0.415 & 22.80 & 0.708 & 0.276\\
    FSGS~\cite{zhu2024fsgs} & \cellcolor[HTML]{FFFFD4}18.80 & \cellcolor[HTML]{FFFFD4}0.531 & \cellcolor[HTML]{FFFFD4}0.418 & \cellcolor[HTML]{FFFFD4}23.28 & \cellcolor[HTML]{FFFFD4}0.715 & \cellcolor[HTML]{FFFFD4}0.274 \\
    CoR-GS~\cite{zhang2024cor} & \cellcolor[HTML]{FFE4CF}19.52 & \cellcolor[HTML]{FFE4CF}0.558 & \cellcolor[HTML]{FFFFD4}0.418 & \cellcolor[HTML]{FFE4CF}23.39 & \cellcolor[HTML]{FFE4CF}0.727 & \cellcolor[HTML]{FFE4CF}0.271 \\
    \textbf{SE-GS} & \cellcolor[HTML]{FFCCC9}19.91 & \cellcolor[HTML]{FFCCC9}0.596 & \cellcolor[HTML]{FFCCC9}0.400 & \cellcolor[HTML]{FFCCC9}23.74 & \cellcolor[HTML]{FFCCC9}0.745 & \cellcolor[HTML]{FFCCC9}0.265 \\
  \bottomrule
  \end{tabular}
  }
  \caption{\textbf{Results on Mip-NeRF360 with 12 and 24 training views.} The first, second, and third-best results are marked in red, orange, and yellow, respectively. }
  \label{tab:360}
\end{table}

\begin{table*}[!t]   
  \centering
  \begin{tabular}{@{}l|ccc|ccc|ccc}
    \toprule
     \multirow{2}{*}{Method}   & \multicolumn{3}{c|}{PSNR$\uparrow$}  & \multicolumn{3}{c|}{SSIM$\uparrow$}   & \multicolumn{3}{c}{LPIPS$\downarrow$} \\
      &  5-view & 7-view & 10-view & 5-view & 7-view & 10-view & 5-view & 7-view & 10-view \\
    \midrule
    3DGS~\cite{kerbl20233d} & 26.48 & 29.06 & \cellcolor[HTML]{FFFFD4}31.82 & 0.819 & 0.878 & \cellcolor[HTML]{FFFFD4}0.924 & \cellcolor[HTML]{FFE4CF}0.301 & \cellcolor[HTML]{FFE4CF}0.228 & \cellcolor[HTML]{FFE4CF}0.184 \\
    FSGS~\cite{zhu2024fsgs} & \cellcolor[HTML]{FFFFD4}26.88 & \cellcolor[HTML]{FFFFD4}29.12 & 31.68 & \cellcolor[HTML]{FFE4CF}0.848 & \cellcolor[HTML]{FFFFD4}0.886 & 0.922 & 0.358 & 0.292 & 0.250  \\
    CoR-GS~\cite{zhang2024cor} & \cellcolor[HTML]{FFE4CF}27.62 & \cellcolor[HTML]{FFE4CF}29.77 & \cellcolor[HTML]{FFE4CF}32.09 & \cellcolor[HTML]{FFFFD4}0.842 & \cellcolor[HTML]{FFE4CF}0.889 & \cellcolor[HTML]{FFE4CF}0.928 & \cellcolor[HTML]{FFFFD4}0.302 & \cellcolor[HTML]{FFFFD4}0.235 & \cellcolor[HTML]{FFFFD4}0.190 \\
    \textbf{SE-GS} & \cellcolor[HTML]{FFCCC9}27.88 & \cellcolor[HTML]{FFCCC9}30.19 & \cellcolor[HTML]{FFCCC9}32.68 & \cellcolor[HTML]{FFCCC9}0.857 & \cellcolor[HTML]{FFCCC9}0.902 & \cellcolor[HTML]{FFCCC9}0.937 & \cellcolor[HTML]{FFCCC9}0.277 & \cellcolor[HTML]{FFCCC9}0.214 & \cellcolor[HTML]{FFCCC9}0.177 \\
  \bottomrule
  \end{tabular}
  
  \caption{\textbf{Results on MVImgNet with 5, 7, and 10 training views.} Results colored in red, orange, and yellow denote the best, second-best, and third-best performances, respectively.}
  \label{tab:mvimgnet}
\end{table*}

\subsection{Quantitative Results}
We provide the quantitative results of the evaluated approaches on the LLFF, DTU, Mip-NeRF360, and MVImgNet datasets in Table~\ref{tab:llff}, Table~\ref{tab:dtu}, Table~\ref{tab:360}, and Table~\ref{tab:mvimgnet}, respectively. The best, second-best, and third-best results in each column are highlighted in red, orange, and yellow, respectively. For MVSplat~\cite{chen2024mvsplat}, we employ the pretrained model released by the authors and evaluate it on DTU. Notably, the per-scene optimized 3DGS methods significantly outperform MVSplat, highlighting the advantage of scene-level radiance field modeling for high-quality novel view synthesis. For these scene-level approaches, the performance of both NeRF and 3DGS methods consistently declines as the number of training views decreases, underscoring the challenge of few-shot novel view synthesis. In this context, our method surpasses all competitors across the four datasets. Note that in sparse-view scenarios, the available information in the training data is limited, making it more challenging to improve performance compared to the dense-view setting. 

Specifically, our method achieves an improvement of $0.34$ in terms of PSNR with 3 views on LLFF. The improvement is larger on DTU, reaching a gain of $0.59$. Since the images in the Mip-NeRF360 dataset are captured from diverse viewpoints, we sample more training views, i.e., 12 and 24, following the setting used in~\cite{zhang2024cor}. On this dataset, our SE-GS achieves the highest PSNR and SSIM, as well as the lowest LPIPS, demonstrating superior NVS quality under the challenges of sparse views with large-scale viewpoint variations. Additionally, to further assess the robustness across diverse scenarios, we conduct an experiment on 50 scenes sampled from the MVImgNet dataset. As reported in Table~\ref{tab:mvimgnet}, our method yields consistently better results than previous state-of-the-art approaches when trained with 5, 7, and 10 views. Our SE-GS also demonstrates better compatibility with relatively dense views. For instance, the PSNR improvement increases from $0.26$ to $0.59$ as the number of training views varies from 5 to 10. 

The methods that leverage auxiliary data terms, such as DNGaussian~\cite{li2024dngaussian} and FSGS~\cite{zhu2024fsgs}, sometimes outperform 3DGS. However, in some cases, such as with 9 views on LLFF, the PSNR of FSGS becomes worse than 3DGS. Since the data acquired through off-the-shelf approaches might be unreliable, the quality of the NVS results is consequently affected. Compared with FSGS and DNGaussian, CoR-GS~\cite{zhang2024cor} exhibits better stability, outperforming 3DGS in more scenarios. This finding highlights the promising potential of regularization in the sparse-view setting. However, the improvement is still not consistent, as CoR-GS shows worse LPIPS than 3DGS when trained with 10 views on the MVImgNet dataset. In contrast, our SE-GS beats 3DGS in all cases, demonstrating superior stability. This is attributed to its ability to effectively aggregate information from diverse perturbed models in the self-ensembling paradigm, leading to robust regularization.

\subsection{Qualitative Results}

\begin{figure}[!t]
    \includegraphics[width=1.0\linewidth]{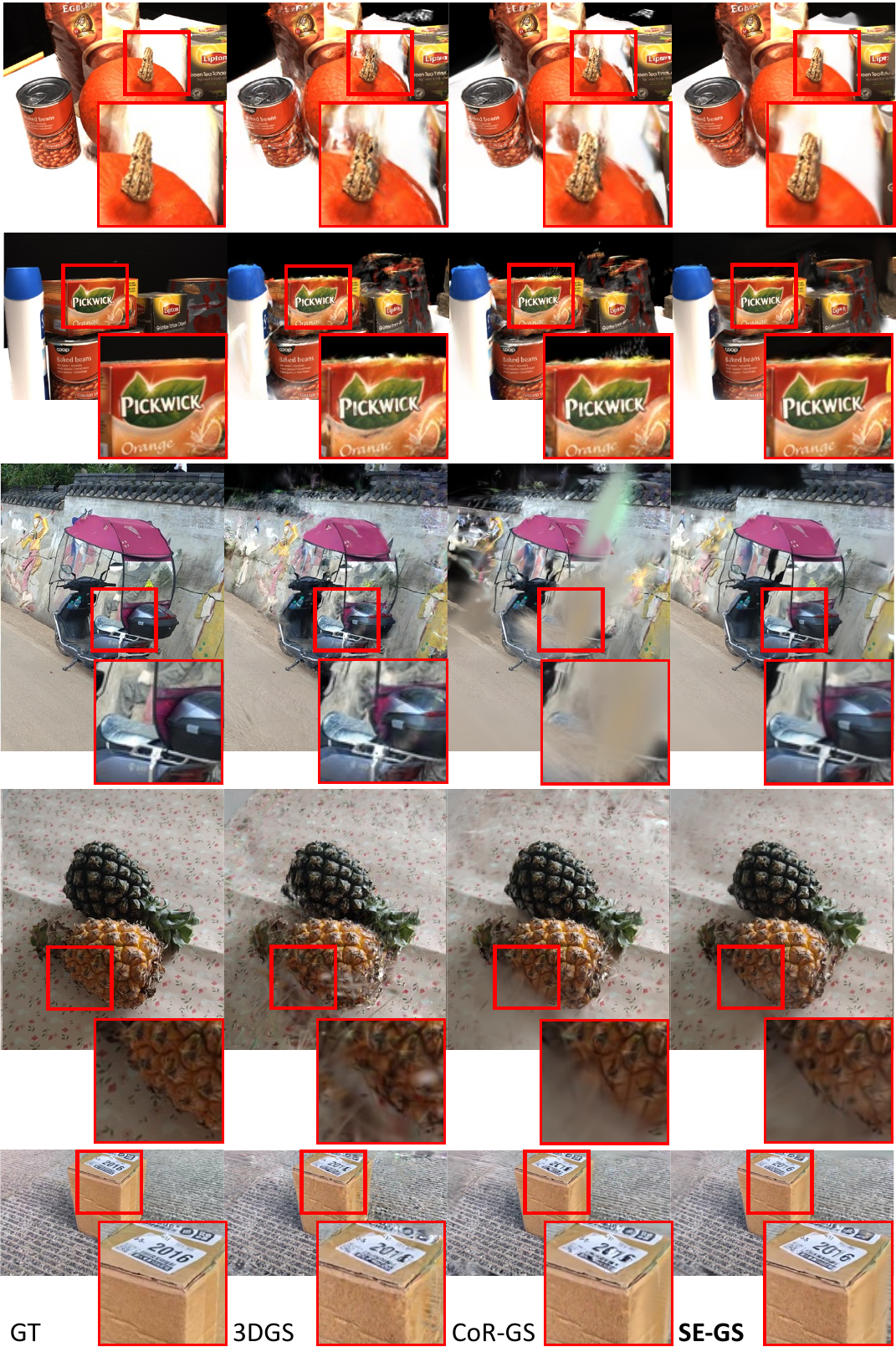}
    \caption{\textbf{Qualitative results.} The methods are trained on sparse views and the renderings of novel views are illustrated. The images are from the DTU and MVImgNet datasets.}
    \label{fig:qual}
\end{figure}

Fig.~\ref{fig:qual} presents qualitative comparisons of 3DGS, CoR-GS, and our SE-GS, showing renderings from novel views. The models are trained on the DTU and MVImgNet datasets with 3 and 5 posed images, respectively. Our SE-GS produces fewer visual artifacts than the other methods, achieving better robustness against the challenge of few-shot training views. Moreover, our method captures finer details, particularly in areas with complex and repeated textures. This demonstrates the effectiveness of our self-ensembling strategy in improving both the visual quality and stability of novel view neural rendering under sparse-view conditions. Please refer to the supplementary material for more visualizations.

\begin{figure}[!t]
    \centering	
    \begin{subfigure}{0.49\linewidth}
        \includegraphics[width=1.0\linewidth]{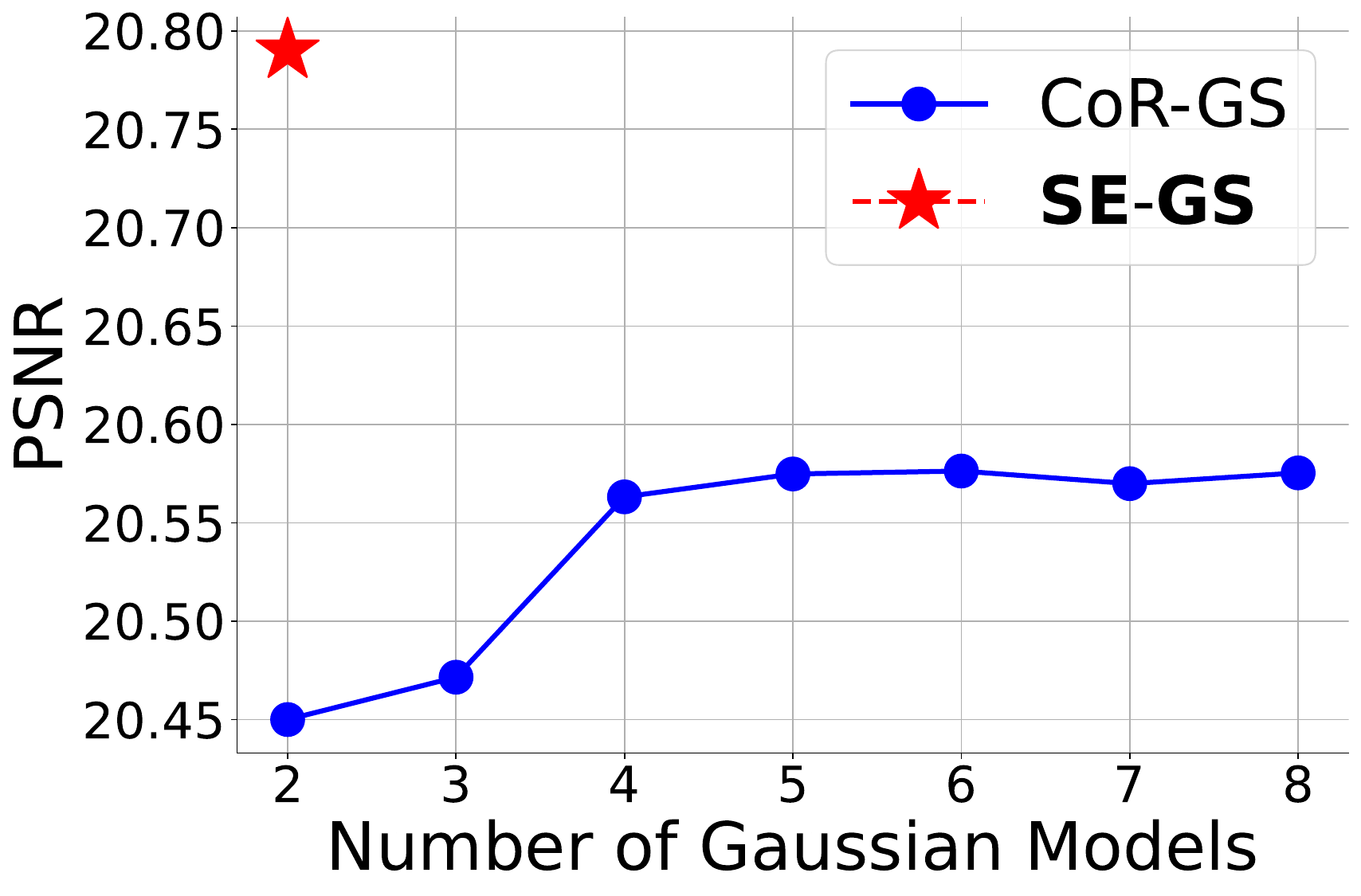}
        \caption{PSNR}
        \label{fig:views_a}
    \end{subfigure}
    \hfill
    \begin{subfigure}{0.49\linewidth}
        \includegraphics[width=1.0\linewidth]{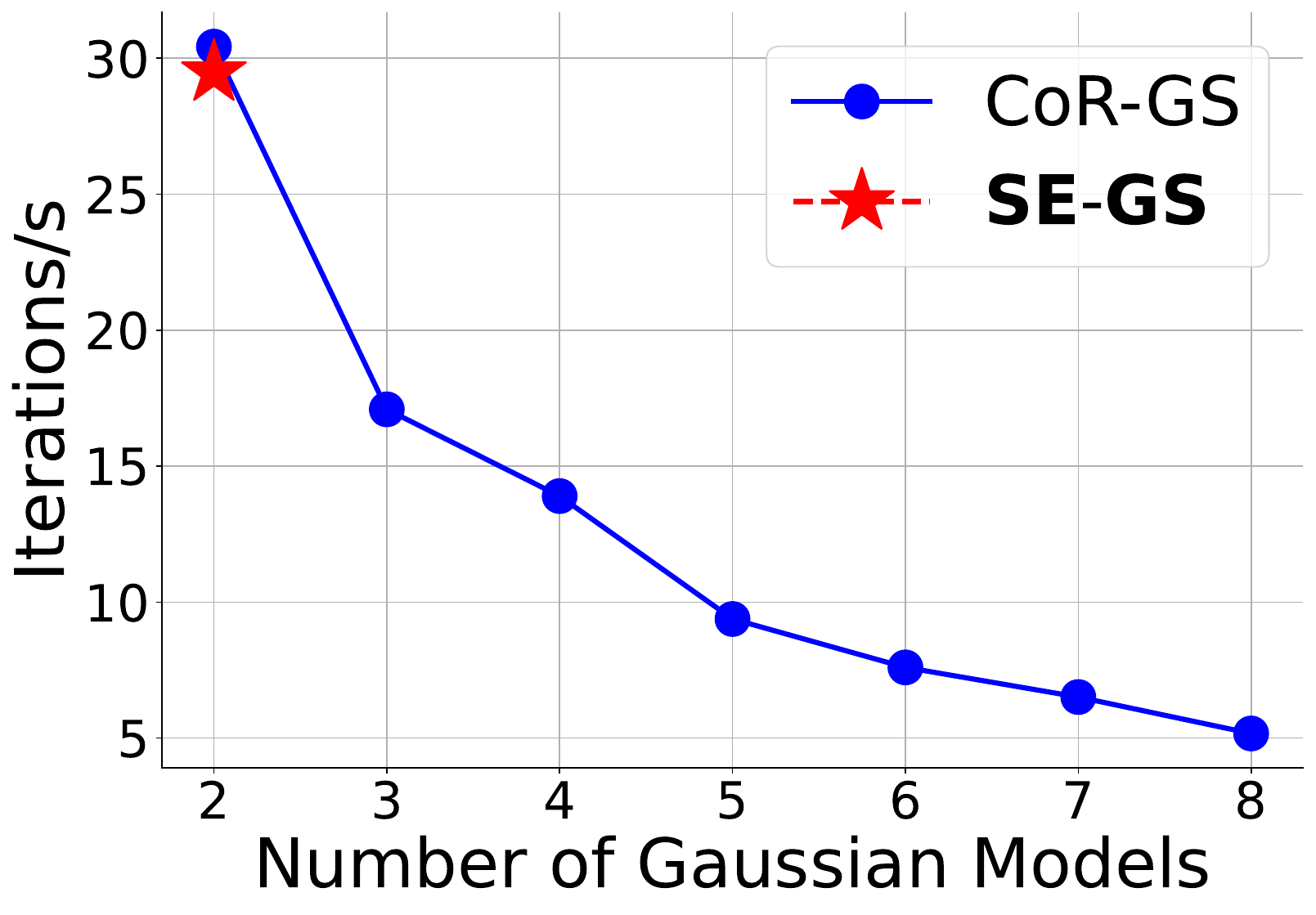}
        \caption{Training Speed}
    \label{fig:views_b}
\end{subfigure}
    \caption{\textbf{Comparison with CoR-GS}. CoR-GS~\cite{zhang2024cor} is trained with different numbers of 3DGS models. The number varies from 2 to 8. The PSNR on LLFF with 3 training views is reported to evaluate the NVS quality. Iterations/s refers to the number of training iterations completed per second, indicating the training speed.} 
    \label{fig:views}
\end{figure}

\subsection{Analysis}

To shed more light on the effectiveness and efficiency of our SE-GS, we perform a comprehensive analysis of the introduced self-ensembling mechanism. The analysis starts by comparing our approach with CoR-GS trained with varying numbers of Gaussian models. The detailed results on LLFF with 3 training views are shown in Fig.~\ref{fig:views}. As shown by Fig.~\ref{fig:views_a}, the PNSR of CoR-GS increases as more Gaussian models are trained. Since adding more Gaussian models enhances the regularization effect in CoR-GS, this observation highlights the potential of improving 3DGS in the sparse-view setting by strengthening the regularization process. The PSNR reaches a peak with 6 Gaussian models, indicating an upper bound of CoR-GS. Our method perturbs the $\mathbf{\Delta}$-model based on the uncertainties computed from dynamically updated image buffers, which results in more diverse 3DGS models compared to CoR-GS. This leads to more effective regularization, as reflected by the better PSNR score. 

Moreover, Fig.~\ref{fig:views_b} shows the training speed measured as the number of training iterations completed per second. The speed of CoR-GS significantly drops as the number of Gaussian models increases. This limitation poses a challenge in scaling CoR-GS up to a large number of Gaussian models. In contrast, in our pipeline, only two models are trained, i.e., the $\mathbf{\Delta}$-model and $\mathbf{\Sigma}$-model. The perturbed models are generated through perturbation. In this context, we achieve a comparable training speed to CoR-GS with 2 models, showing that the perturbation process incurs a negligible additional cost. Note that only the $\mathbf{\Sigma}$-model is retained for NVS during inference, so the testing speed of SE-GS is the same as that of the original 3DGS. Consequently, our method is capable of efficiently enhancing the performance of 3DGS with sparse training views.

\begin{table}[!t]   
  \centering
  \resizebox{1.0\linewidth}{!}{
  \begin{tabular}{@{}cccc}
    \toprule
    Method & Random & Gradient-Aware & \textbf{Uncertainty-Aware} \\
    \hline 
    PSNR$\uparrow$ & 18.63 & 19.18 & 20.34 \\
    \bottomrule
  \end{tabular}
  }
  \caption{\textbf{Effectiveness of the uncertainty-aware perturbations.} Random indicates that we add random perturbations to all Gaussians in the $\mathbf{\Delta}$-model. Gradient-Aware denotes an alternative in which we perturb the Gaussians based on gradients. All methods are trained on LLFF with 3 views for 2000 iterations, and PSNR is utilized as the metric.}
  \label{tab:noise}
\end{table}

\begin{figure}[!t]
    \centering	
    \includegraphics[width=0.7\linewidth]{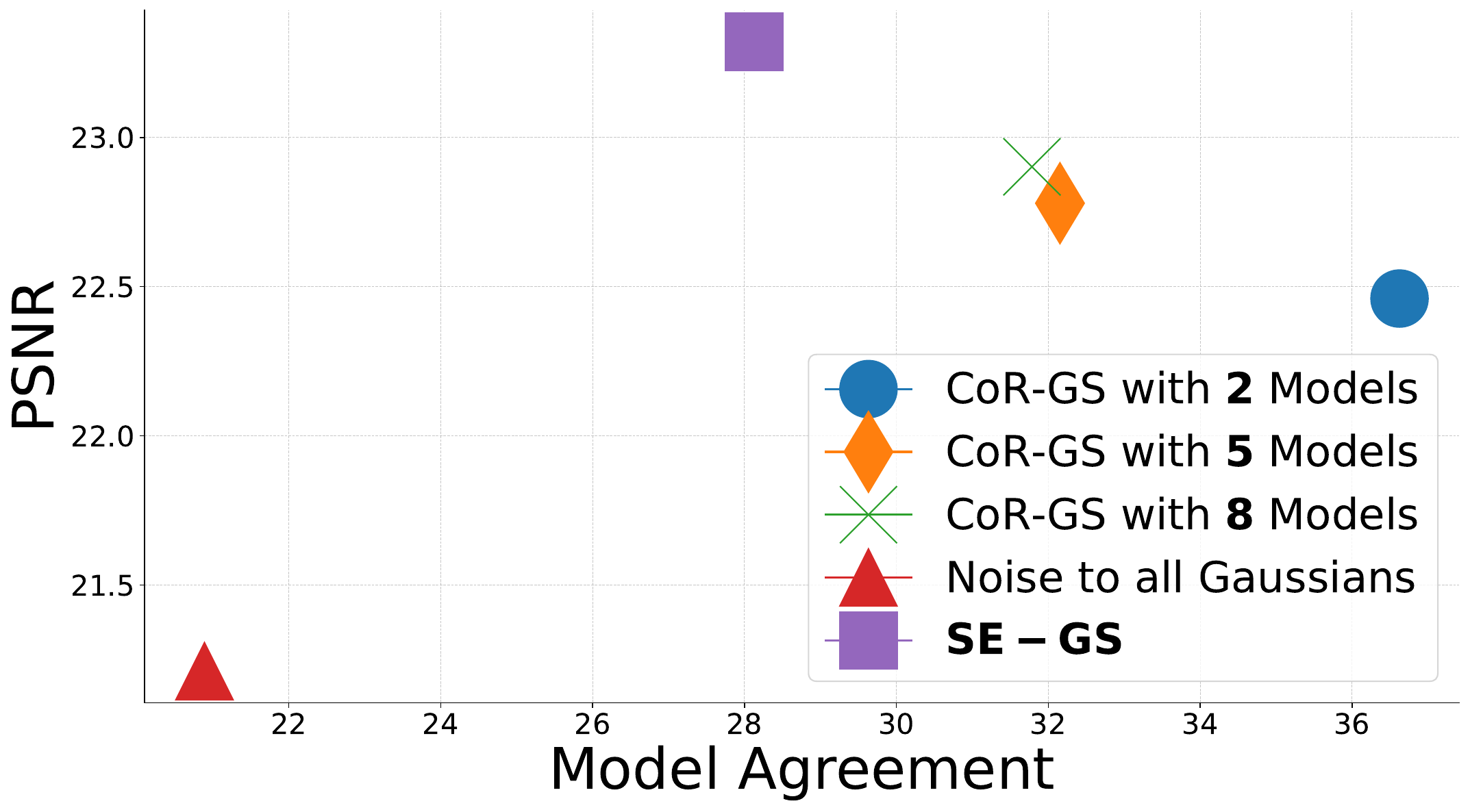}
    \caption{\textbf{Impact of model diversity.} The $x$-axis represents model agreement, indicating the similarity among models involved during training.  Higher agreement values suggest greater model similarity and lower diversity. The $y$-axis denotes the PSNR of the methods during testing. The experiment is conducted on LLFF with three training views.}
    \label{fig:aba_ensemble}
\end{figure}

Finally, we assess the effectiveness of our uncertainty-aware perturbation strategy, comparing it with two alternatives. As introduced in Sec.~\ref{sec:method}, a straightforward method for perturbing the $\mathbf{\Delta}$-model is to add random noise to the parameters of all Gaussians. We denote this approach as \emph{Random}. Building upon this baseline, an alternative approach, referred to as \emph{Gradient-Aware}, identifies unreliable Gaussians by analyzing the gradients. Specifically, we replace the indicator function in Eq.~\ref{eq:indicator} with one based on gradient magnitudes. Gaussians with gradient magnitudes exceeding a threshold are perturbed. In practice, we adopt the same threshold as used in density control. We train these two alternatives and our approaches on LLFF with 3 views for 2000 iterations, which challenges the methods in terms of convergence speed. As listed in Table~\ref{tab:noise}, \emph{Random} performs the worst, indicating that this strategy negatively impacts convergence speed. Notably, our uncertainty-aware strategy aggregates information from multiple training steps, while the gradient-aware method only utilizes the gradients at the current training step. The effectiveness of this design is evidenced by the improvement in PSNR shown in Table~\ref{tab:noise}. Additionally, to further investigate the impact of model diversity, we conduct a detailed analysis on LLFF. As shown in Fig.~\ref{fig:aba_ensemble}, the $x$-axis denotes the model agreement measured as PSNR between renderings of all involved models on the test views; the $y$-axis shows PSNR between the NVS result and the ground-truth image during testing. The Model agreement reflects consistency, where higher values indicate greater similarity among models. The agreements of CoR-GS with 5 and 8 models are nearly identical, meaning that the additional models lack sufficient diversity to enhance the ensembling process. This aligns with the observation in Fig.~\ref{fig:views_a} that merely using more models yields limited improvement. While perturbation can enhance model diversity, it may weaken supervision reliability. As shown by the red triangle in Fig.~\ref{fig:aba_ensemble}, adding noise to \emph{all} Gaussians yields overly diverse models. The supervision in the regularization term becomes too noisy, resulting in limited PSNR. In contrast, our uncertainty-aware perturbation method only perturbs unreliable Gaussians, enabling a good trade-off between \emph{model diversity} and \emph{supervision reliability}.

%% file: sec/5_conclusion_arxiv.tex
\section{Conclusion}
\label{sec:conclusion}
In this paper, we have presented a new self-ensembling mechanism that enhances the novel view synthesis of 3DGS with sparse training images. We have tackled the challenge of overfitting by training an ensemble named $\mathbf{\Sigma}$-model with the guidance of diverse perturbed models derived from a $\mathbf{\Delta}$-model. To obtain such models, we have introduced an effective strategy that perturbs the $\mathbf{\Delta}$-model based on uncertainties computed from dynamically updated buffers of pseudo-view renderings. We have conducted experiments on LLFF, DTU, Mip-NeRF360, and MVImgNet, as well as a comprehensive analysis of our approach. The experimental results have demonstrated the effectiveness, stability, and efficiency of our SE-GS. In future work, we plan to incorporate an outlier identifier into our perturbation mechanism to reduce the impact of unreliable models in regularization and facilitate the self-ensembling process.
\\

\noindent\textbf{Acknowledgment.} This work was funded in part by the Swiss National Science Foundation via the Sinergia grant CRSII5-180359 and the Swiss Innovation Agency (Innosuisse) via the BRIDGE Discovery grant 40B2-0\_194729.

%% file: sec/X_suppl.tex
\clearpage
\setcounter{page}{1}
\maketitlesupplementary

\section*{Details on Ensembling Learning}
\begin{figure}[!t]
    \centering	
    \includegraphics[width=1.0\linewidth]{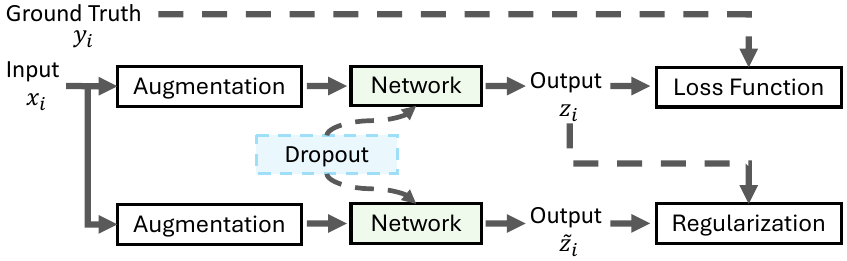}
    \caption{\textbf{Pipeline of self-ensembling.} Network variants are generated via dropout, and self-ensembling is achieved by utilizing a regularization term between the variants.}  
    \label{fig:apendix_ensemble}
\end{figure}
Ensemble learning has been recognized for its effectiveness in improving robustness and generalization~\cite{ganaie2022ensemble}, especially in scenarios with limited training data. We illustrate the training pipeline of existing self-ensembling approaches~\cite{french2017self} in Fig.~\ref{fig:apendix_ensemble}. Specifically, self-ensembling aims to aggregate information of diverse models without training multiple models from scratch. The model diversity is typically enhanced via dropout. In this context, the variants are dynamically derived from a single network during training, thereby prompting both diversity and efficiency. Self-ensembling is then achieved by incorporating a regularization term, which operates alongside the task-specific loss function based on ground truth. This regularization enforces consistency among the network variants by constraining their outputs, thereby mitigating model bias and enhancing robustness and generalization. Inspired by its effectiveness, we introduce the first self-ensembling Gaussian Splatting method to improve novel view synthesis in scenarios with sparse training views.

\section*{Details on Uncertainty-Aware Perturbation}
Recall that in Eq.~\ref{eq:indicator} of the main paper, we associate the reliability of Gaussians with the pixel-level uncertainty scores. In practice, for each uncertainty map $\hat{\mathbf{U}}$, we define $\tau$ as
\begin{align}
\label{eq:thr}
\tau &= \text{max}(\hat{\mathbf{U}}_{\text{sorted}} \left[ \left\lceil r * |\hat{\mathbf{U}}_{\text{sorted}}| \right\rceil \right], \theta),
\end{align}
where $\hat{\mathbf{U}}_{\text{sorted}}$ is obtained by sorting $\hat{\mathbf{U}}$ in descending order, $|\hat{\mathbf{U}}_{\text{sorted}}|$ denotes the number of uncertainty scores in the map, $\left\lceil \cdot \right\rceil$ indicates the ceiling function, $\hat{\mathbf{U}}_{\text{sorted}} \left[ \cdot \right]$ accesses the value at the specified index. Here, $r=0.05$ is a ratio scalar, and $\theta=0.01$ serves as a minimum tolerance. Subsequently, we identify the Gaussians that overlap with the pixels having uncertainty scores greater than $\tau$. For each Gaussian in the $\mathbf{\Delta}$-model, we splat it to the 2D image planes corresponding to uncertainty maps. If any of these pixels are covered by the 2D splats of this Gaussian, the indicator function in Eq.~\ref{eq:indicator} returns 1, marking the Gaussian as unreliable.

In Eq.~\ref{eq:perturb_ours}, we sample the noise $\delta_{t}$ from a normal distribution. As an example, we elaborate on the perturbation process for the 3D position, which is similar for the other Gaussian parameters. In this context, we define $\delta_{t}$ as $\delta_{t}\in\mathcal{N}(\mathbf{0}, \sigma^{2}\mathbf{I})$ where $\mathbf{I}\in\mathbb{R}^{3\times 3}$ is an identity matrix. In practice, we adaptively adjust $\sigma$ based on the magnitude of the parameters, as formulated by
\begin{align}
\label{eq:sigma}
\sigma=\omega\frac{1}{N}\sum_{i=1}^{N}{||\mathbf{\mu}_i||}_1,
\end{align}
where $\omega$ is a scalar that controls the noise level and ${||\cdot||}_1$ denotes the L1 norm of the Gaussian parameter. The value of $\omega$ decays from 0.08 to 0.02, following a decay function introduced in~\cite{fridovich2022plenoxels}.

\section*{Setup on DTU}
As we mentioned in the main paper, COLMAP fails in some scenes on DTU when using sparse-view images. In our experiments, we randomly initialize the point cloud in these scenes. Specifically, we use random initialization for \emph{scan8}, \emph{scan40}, and \emph{scan110} with 3 training views, and for \emph{scan21} with 6 training views.

\section*{Ablation Studies}
To shed more light on the impact of noise during the perturbation, we conduct comprehensive ablation studies on the perturbation interval and noise level. 

\begin{figure}[!t]
    \centering	
    \includegraphics[width=0.7\linewidth]{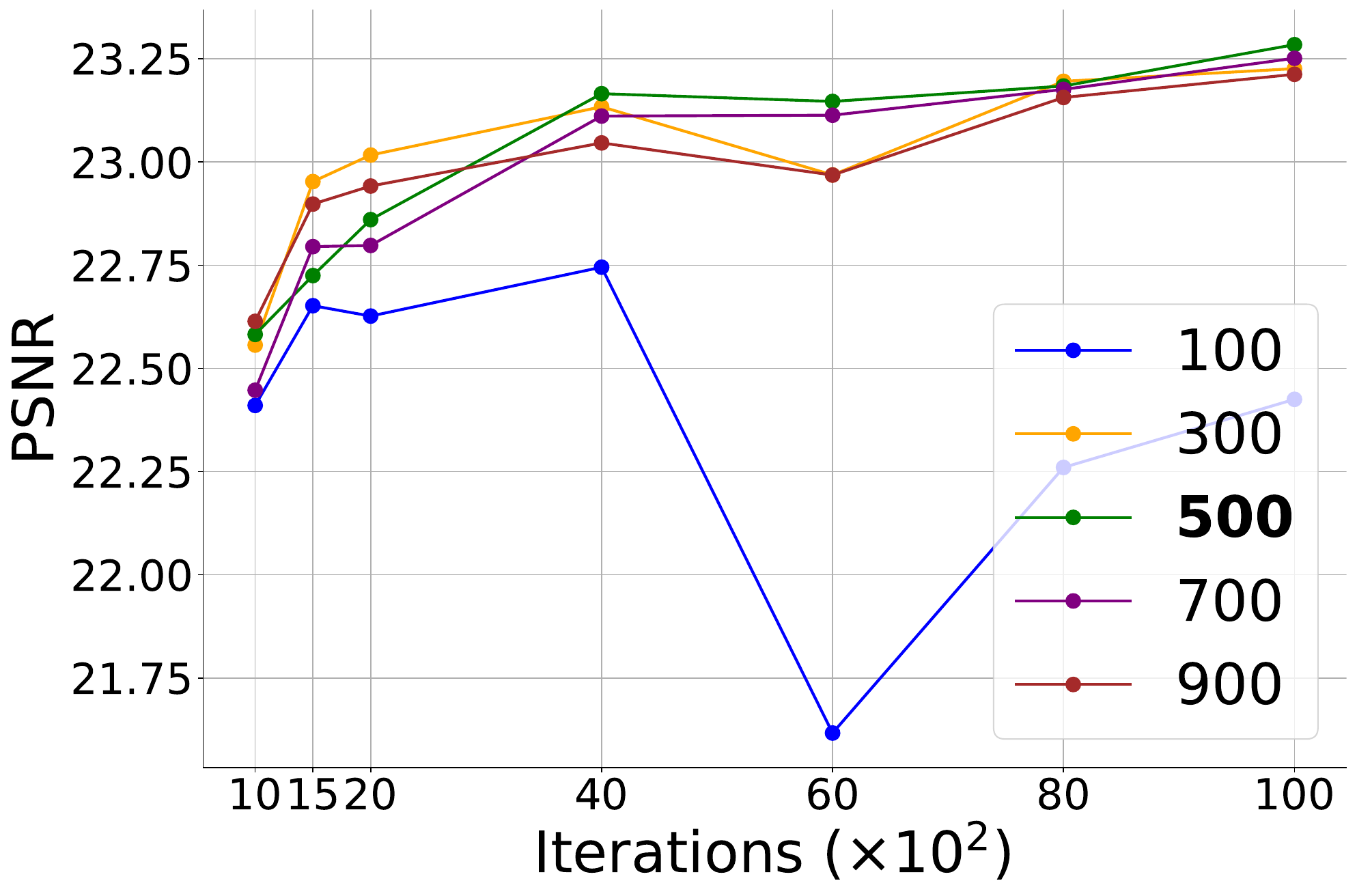}
    \caption{\textbf{Ablation study on the perturbation interval.} PSNR values on testing views throughout training are reported. The interval varies from 100 to 900.}  
    \label{fig:aba_interval}
\end{figure}

\begin{figure}[!t]
    \centering	
    \includegraphics[width=0.7\linewidth]{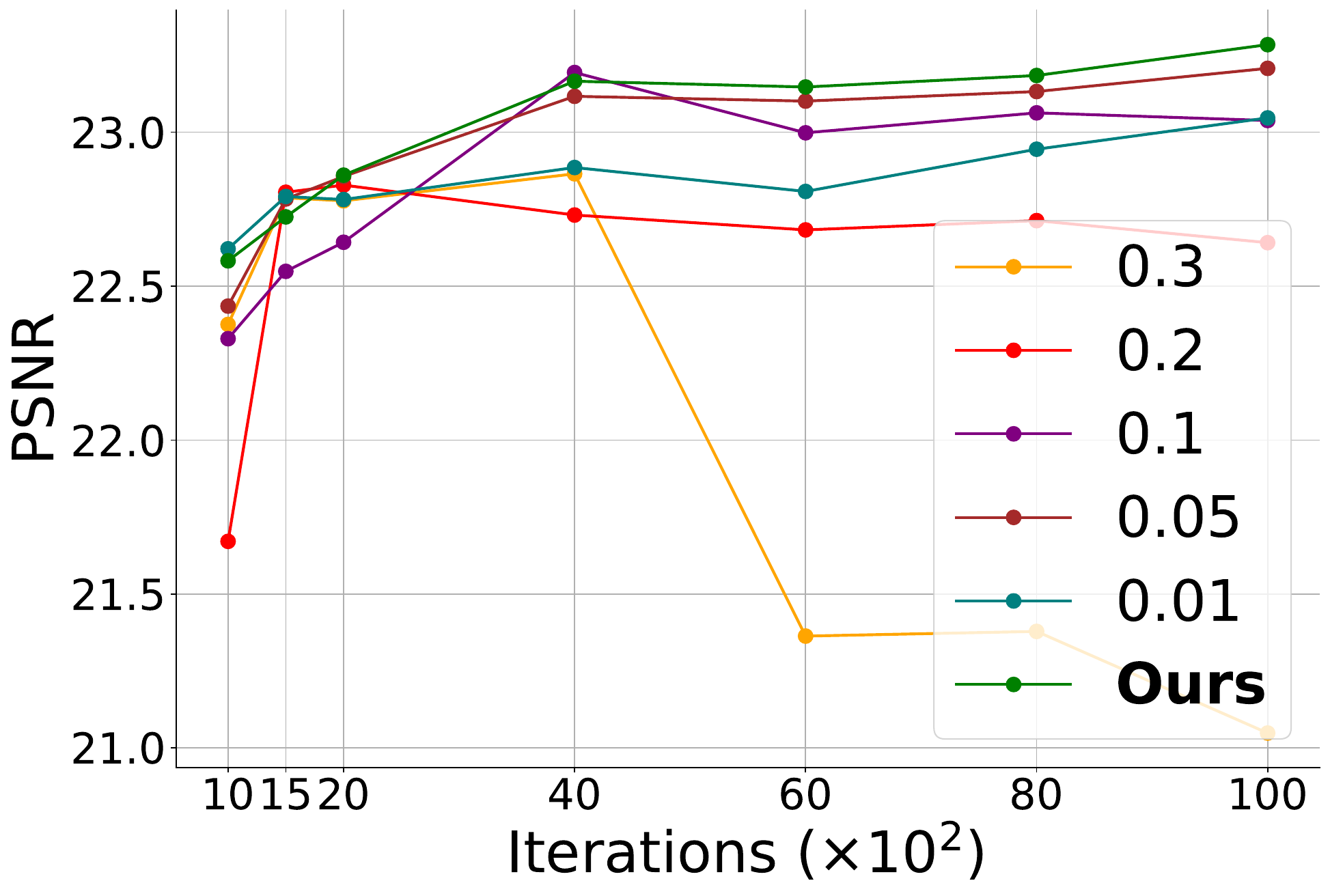}
    \caption{\textbf{Ablation study on the noise level.} We evaluate our method using different values of $\omega$, each corresponding to a distinct noise level.} 
    \label{fig:aba_sigma}
\end{figure}

In our experiments, we perturb the $\mathbf{\Delta}$-model every 500 iterations by default. To evaluate the effect of different intervals, we test our method on the LLFF dataset with 3 training views, using intervals ranging from 100 to 900. As shown in Fig.~\ref{fig:aba_interval}, the curves represent the PSNR values obtained on testing views at different training iterations. Perturbing the $\mathbf{\Delta}$-model too frequently, e.g., with an interval of 100, significantly degrades performance, indicating a negative impact on our method. Conversely, large intervals, such as 900, reduce the number of perturbed models, thereby weakening the self-ensembling effect and leading to decreased performance compared to the default setting.

Moreover, we analyze the effect of the noise level by conducting experiments with varying values of $\sigma$ in Eq.~\ref{eq:sigma}. Specifically, we adjust the value of $\omega$ from 0.01 to 0.3, corresponding to increasing noise levels. We train our method with each specific noise level on the LLFF dataset using 3 training views. The results on testing views at different training iterations are shown in Fig.~\ref{fig:aba_sigma}. Strong perturbation, such as those with $\omega$ values of 0.3 and 0.2, result in significantly worse performance compared to other settings. In these cases, the perturbed models are too far from the $\mathbf{\Delta}$-model in the Gaussian parameter space, reducing the consistency among the variants. The performance is also limited when a small $\omega$, such as 0.01, is used. In this scenario, all perturbed models closely resemble the $\mathbf{\Delta}$-model, diminishing the benefits of self-ensembling.
In contrast to using a fixed $\omega$, we adopt a decay function, where $\omega$ dynamically varies from 0.08 to 0.02 during training. This strategy achieves a good trade-off between the consistency and diversity of the perturbed models.

\begin{figure}[!t]
    \centering	
    \includegraphics[width=0.7\linewidth]{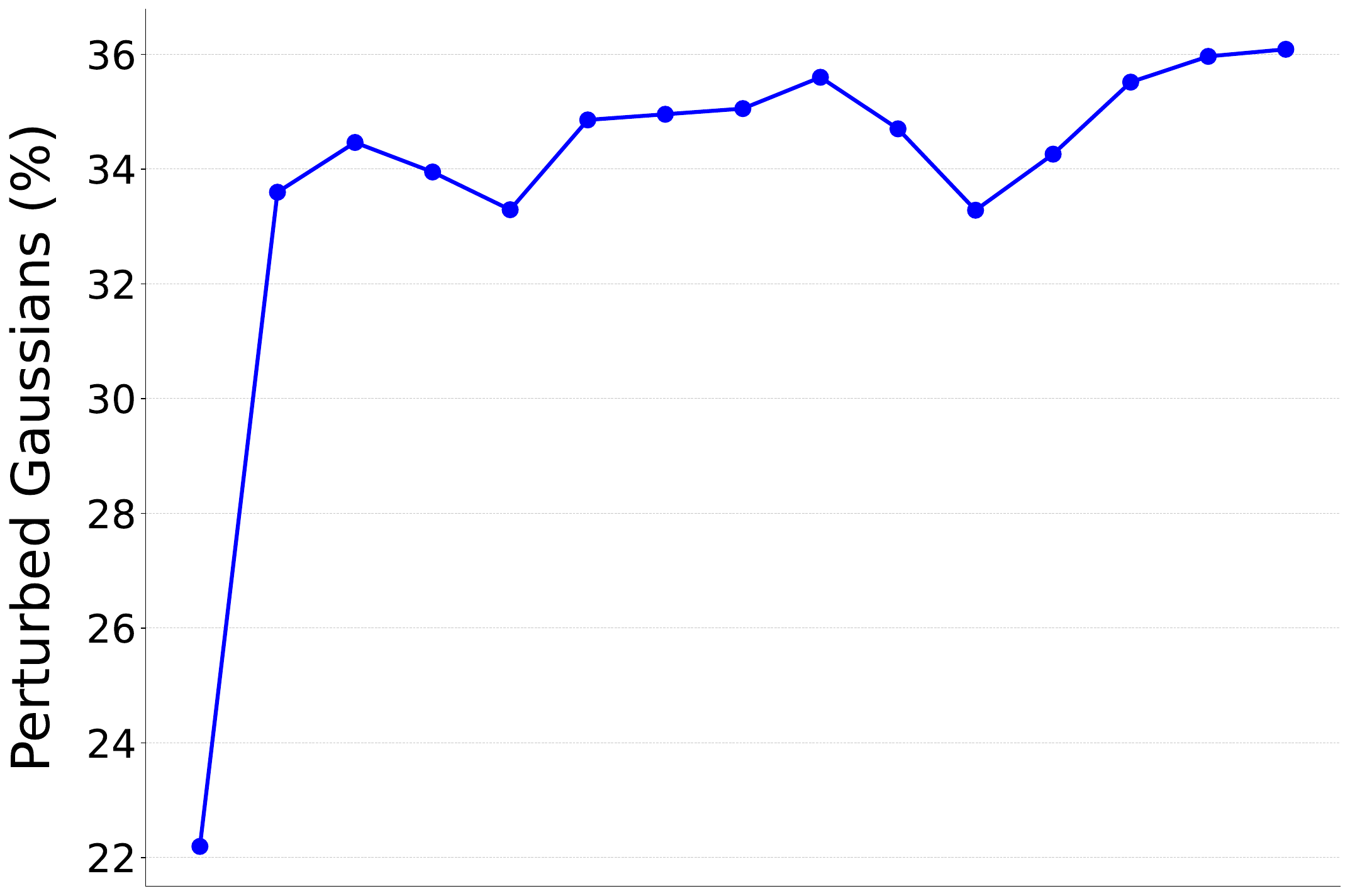}
    \caption{\textbf{Number of perturbed Gaussians throughout training.} The curve illustrates the percentage of perturbed Gaussians in the $\mathbf{\Delta}$-model at different training iterations.} 
    \label{fig:aba_num}
\end{figure}

To better understand the presented perturbation mechanism, in Fig.~\ref{fig:aba_num}, we illustrate the percentage of perturbed Gaussians in the $\mathbf{\Delta}$-model. Note that the threshold $\tau$ in Eq.~\ref{eq:thr} is defined adaptively. Therefore, the percentage dynamically varies throughout training. This dynamic behavior balances the diversity and reliability of the perturbed models, thereby enhancing the self-ensembling process.

\section*{More Visualization Results}
We provide more visualizations including pseudo-view renderings of the $\mathbf{\Sigma}$-model and the $\mathbf{\Delta}$-model, uncertainty maps, and NVS results. For more details, please refer to the uploaded videos. \emph{nvs\_res.mov}: Shows the novel-view renderings of 3DGS, CoR-GS, and our SE-GS, where our method results in finer details and fewer artifacts. \emph{uncertainty\_map\_update.mov}: Records the update of uncertainty maps during training. For each scene, we sample a pseudo view and store the corresponding rendering of the $\mathbf{\Delta}$-model in the image buffer. The uncertainty map is then derived from this buffer. The updates across different training iterations are shown in the video. \emph{sigma\_model\_vs\_perturbed\_model.mov}: Displays the pseudo-view renderings of the $\mathbf{\Sigma}$-model and the perturbed $\mathbf{\Delta}$-model used in the regularization Eq.~\ref{eq:reg}.

%%% visualization of uncertainty maps and perturbed images
\clearpage